\documentclass[acmsmall]{acmart}
\usepackage[squaren]{SIunits}
\usepackage{tabularx}
\usepackage{array}
\usepackage{enumitem}
\usepackage{bm}
\newcolumntype{P}[1]{>{\centering\arraybackslash}p{#1}}

\AtBeginDocument{%
  \providecommand\BibTeX{{%
    \normalfont B\kern-0.5em{\scshape i\kern-0.25em b}\kern-0.8em\TeX}}}

\setcopyright{acmlicensed}
\acmJournal{PACMCGIT}
\acmYear{2023} \acmVolume{6} \acmNumber{2} \acmArticle{1} \acmMonth{8} \acmPrice{15.00}\acmDOI{10.1145/3606931}

\acmSubmissionID{5151}

\newcommand{\zhaoming}{}

\setcitestyle{nosort}

\begin{document}


\title{Hierarchical Planning and Control for Box Loco-Manipulation}



\author{Zhaoming Xie}
\email{zhxie@stanford.edu}
\affiliation{%
  \institution{Stanford University}
  \country{USA}
}

\author{Jonathan Tseng}
\email{jtseng20@cs.stanford.edu}
\affiliation{%
  \institution{Stanford University}
  \country{USA}}

\author{Sebastian Starke}
\email{sstarke@fb.com}
\affiliation{%
 \institution{Meta}
 \country{United Kingdom}}

\author{Michiel van de Panne}
\email{van@cs.ubc.ca}
\affiliation{%
 \institution{University of British Columbia}
 \country{Canada}}

\author{C. Karen Liu}
\email{karenliu@cs.stanford.edu}
\affiliation{%
 \institution{Stanford University}
 \country{USA}}


\begin{abstract}
Humans perform everyday tasks using a combination of locomotion and manipulation skills. Building a system that can handle both skills is essential to creating virtual humans. We present a physically-simulated human capable of solving box rearrangement tasks, which requires a combination of both skills. We propose a hierarchical control architecture, where each level solves the task at a different level of abstraction, and the result is a physics-based simulated virtual human capable of rearranging boxes in a cluttered environment. The control architecture integrates a planner, diffusion models, and physics-based motion imitation of sparse motion clips using deep reinforcement learning. Boxes can vary in size, weight, shape, and placement height. Code and trained control policies are provided.
 \end{abstract}

\ccsdesc[300]{Computing methodologies~Physical simulation}
\ccsdesc[300]{Computing methodologies~Reinforcement learning}

\keywords{character animation, loco-manipulation}

\begin{teaserfigure}
  \includegraphics[width=0.8\textwidth]{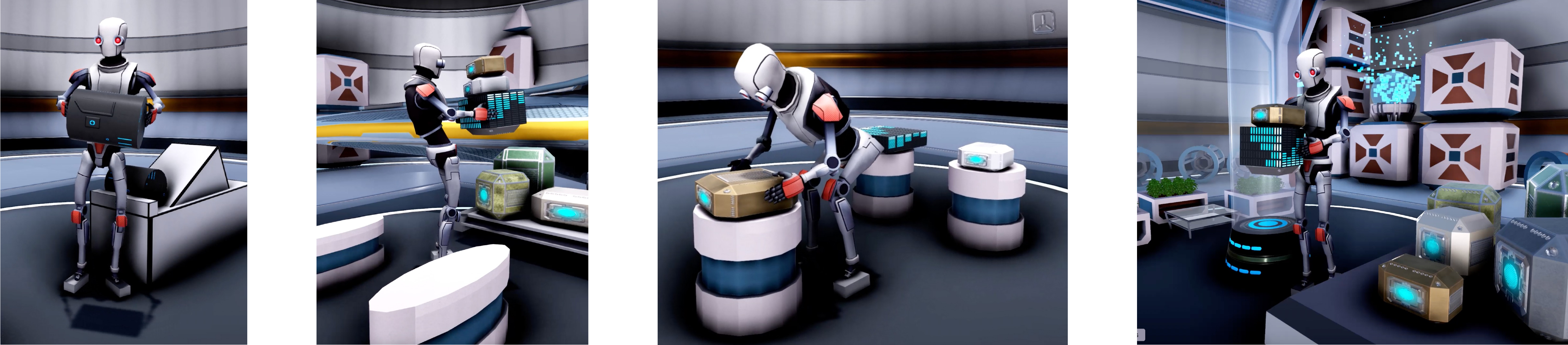}
  \centering
  \caption{We develop loco-manipulation skills for box-carrying physics-based characters.
  This is achieved via a high-level A* planner, a mid-level diffusion-based motion, 
  and a low-level RL control policy.
  }
  \label{fig:teaser}
\end{teaserfigure}

\maketitle

\section{Introduction}


Organizing our living space is a routine and familiar part of the human experience -- from a young age, we are taught to “put things back in their place”.
However, these types of effortless tasks, e.g., object (re)arrangement in a cluttered environment, remain challenging to synthesize in computer animation, as the character must be capable of navigating around obstacles, locomotion, and manipulation in arbitrary, diverse settings.
We present a physically-simulated human model capable of autonomously performing this type of task,
demonstrating the ability to lift, carry, and place boxes in environments with obstacles.
As with other recent efforts, e.g., ~\cite{2019-TOG-neural_state_machine,2020-catch_and_carry,zhang2022couch}, our goal is to 
advance the ability of characters to use locomotion and manipulation to interact with their environments in meaningful ways.

We adopt a hierarchical approach to solving these rearrangement tasks.
This decouples the high-level spatial planning from low-level motor-skill execution and therefore allows for generalization of the locomotion and manipulation skills to a wide range of arrangement tasks. 
At the highest level, we use a simple planner to establish a basic path between the pick-up and place-down locations. 

The middle level is responsible for generating realistic kinematic locomotion trajectories.
While there exist many techniques for locomotion synthesis, we employ a diffusion model for its simplicity and flexibility
and the ready availability of walking motions in human motion datasets. We investigate its potential for synthesizing realistic human locomotion from high-level path constraints. We introduce a \emph{bidirectional root} representation to improve the diffusion model's ability to satisfy waypoint conditions more precisely. However,  motion data for box pickup and placement is in shorter supply, and so here we use a single motion clip as a kinematic reference. 

Lastly, low-level policies provide fully physically simulated motor control skills capable of imitating the kinematic reference motions.  This is accomplished using an imitation-based deep-reinforcement learning policy that is further made object-aware. This allows for robust carrying behaviors which can accommodate a variety of box weights, sizes, and pick-up or place-down heights. 

We evaluate our method and its ability to generalize by demonstrating its capabilities in various scenarios.
We demonstrate the robustness of locomotion and manipulation components separately and conduct ablation studies.

Our primary contributions are as follows:
(a) a hierarchical planning-and-control system that endows characters with box locomanipulation skills;
(b) the introduction of diffusion models with bidirectional root control as a suitable planning method for physics-based control; and
(c) the ability to generalize a single box lift-and-place motion to physics-based
    lift-and-place skills that span a range of box weights and heights.


\section{Related Work}

\begin{figure*}[t]
    \centering   \includegraphics[width=0.9\columnwidth]{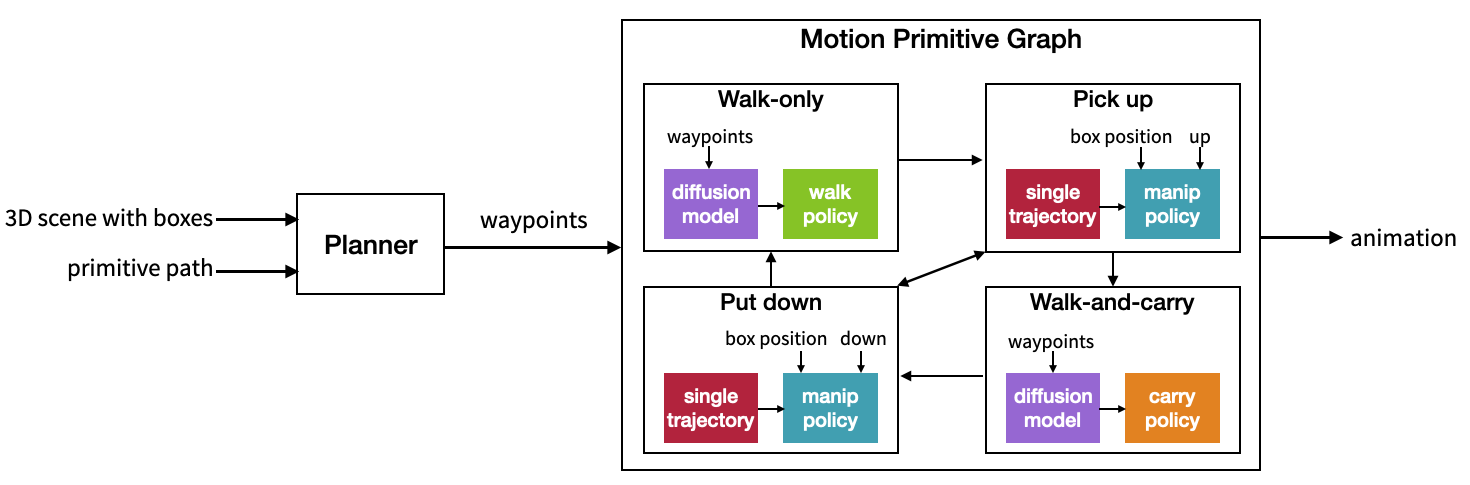}
    \caption{System overview. We design four motion primitives for locomotion and manipulation which can be combined to accomplish arrangement tasks in a 3D scene according to a simple motion primitive graph. Each primitive consists of a policy that controls the character in a physically simulated environment. The locomotion primitives, ``Walk-only'' and ``Walk-and-carry'', leverage a diffusion model to generate kinematic reference trajectories that guide the policies. Components with the same color are identical.  
    }
    \label{fig:system}
\end{figure*}

Realistic animated motion can be generated using kinematic methods or using physics-based simulations.  
The problem is difficult enough that it is common to
focus on making progress on specific subtasks, such as locomotion \cite{2017-siggraph-deeploco, 2007-simbicon, 2022-control-vae, 2017-siggraph-pfnn} and 
manipulation \cite{liu2009dextrous, 2022-chopstick, 2021-dash, 2021-manipnet}. 
However, humans can fluently sequence and combine locomotion and manipulation with an ease
that is challenging to reproduce.



\subsection{Loco-manipulation for Humanoids}
Several methods have been proposed to combine locomotion and manipulation using kinematics-based approaches, e.g., \cite{2012-concurrent_manipulation,2016-task-based-locomotion, 2019-TOG-neural_state_machine, 2022-cvpr-goal}. Kinematic-based approaches can generate high-quality character motion but can still
lack detail and weight that comes from the physical nature of the interaction between objects and between the character and the object.

Physics-based simulation of loco-manipulation behavior provides a principled way to address these issues, but comes with its own challenges.
The interaction with the objects is often simplified, e.g.,\cite{2012-contact_invariant, coros2010generalized} or requires significant compute resources and 
may still be limited in terms of motion quality, e.g., \cite{2020-catch_and_carry}.
Learned behaviors for approaching a chair and sitting are demonstrated in ~\cite{chao2021learning}.
This is accomplished via a set of walk, turn, and sit controllers, that imitate motion clips together with an additional navigation-target task reward.
A higher-level controller is then trained which both selects a given lower-level controller and its target-action. 
By comparison, we demonstrate the ability to avoid obstacles, physically track diverse motions from diffusion-model trajectories, 
and perform richer interaction with objects, e.g., moving them.
Another recent approach demonstrates how to use a physics-based RL controller as a projection-step during the diffusion to obtain
more physically-plausible results from diffusion models~\cite{yuan2022physdiff}, driven by language-based input.  
We show that diffusion models can generate sufficiently good plans to directly track with RL-based controllers, 
demonstrate control without residual forces, and control tasks involving planning and environment interaction. \zhaoming{Concurrent work ~\cite{synthesizing_interaction} also demonstrates full-body object rearrangement tasks using the adversarial motion prior framework. Our hierarchical framework allows for manipulating multiple boxes simultaneously and navigation in more diverse environments.}

Loco-manipulation for humanoids is also an important topic in robotics, where prior work demonstrates whole-body control strategies for picking and carrying a single object, \cite{2005-humanoid_carrying, 2007-humanoid-lifting}. 
The control needed to carry a stack of boxes has also been demonstrated on a humanoid robot, e.g., \cite{2021-carrying_multiple}.
The locomotion behavior is often slow or quasi-static and heavily engineered with model-based methods. 
In our work we seek to develop largely model-free methods that instead rely on recent data-driven kinematic methods in combination with 
reinforcement learning for physics-based control.

\subsection{Reinforcement Learning for Character Animation}
Reinforcement learning (RL) has become a popular framework to generate character animations~\cite{kwiatkowski2022survey}, including  with physics-based simulation. Prior work use RL to generate diverse skills for virtual humans, e.g., \cite{2018-siggraph-deepmimic, 2019-siggraph-scadiver, jumping, 2018-SIGGRAPH-basketball}. Diverse motion sequence can be generated by combining a motion generator and a control policy that is trained to imitate a large range of motion capture data, e.g., \cite{2019-siggraph-drecon, 2019-SiggraphAsia-basketball}. They often require users to provide input such as moving direction and speed to complete a task. Hierarchical architectures are proposed such that an RL policy can automatically complete high-level tasks by utilizing the low-level motor skills, e.g., \cite{2022-TOG-ASE, 2020-catch_and_carry, 2022-control-vae, 2020-siggraph-motion_vae, 2022-siggraph-asia-control_vae, 2017-siggraph-deeploco}.
However, they are restricted in their ability to manipulate their environments.

\subsection{Kinematic Motion Generation for Character Animation}
Kinematic-based motion synthesis allows for high quality motion generation with the aid of motion capture data. 
Most progress has been made on modeling movement whose principal interactions with the environment are that of the feet contacting the ground,
such as the many styles and speeds of locomotion, dancing, or isolated sports motions.
Motion capture clips can be queried based on user input, using techniques such as motion graph  \cite{kovar2008motion} or motion matching \cite{2018-motion_matching}. Supervised learning based techniques allow for similar capabilities with limited motion capture data, e.g., \cite{2017-siggraph-pfnn, RNN-motion-generation}. Generative models can be used to synthesize diverse motions given the same input, e.g., variational autoencoders \cite{2020-siggraph-motion_vae, 2021-ICCV-Humor}, and, most recently, diffusion models \cite{2022-motion_diffusion, dance_diffusion}.

\section{System Overview}
Our system aims to generate animations of a physically simulated character arranging the boxes by picking them up, carrying them, and putting them down while navigating the 3D scene to achieve the desired configuration for those boxes. Figure \ref{fig:system} shows an overview of our system. 

Rearrangement tasks in a contextual environment require coordinated locomotion and manipulation which can be broken down to four primitive motions: walk-only without objects, walk-and-carry objects, pick-up objects, and put-down objects. We construct a simple graph to express the sequential relationships among those primitives: an arrow from primitive A to B indicates that B can be executed after A but not vice versa (Figure \ref{fig:system}, middle).


Given a 3D scene and a sequence of primitives (primitive path) that respects the primitive graph, a high-level planner generates a sequence of waypoints (root positions and heading directions in the ground plane) for each locomotion primitive.
This ensures that the character avoids collisions in the scene while navigating to desired locations to complete a sequence of pick-up and put-down tasks.

Once planned, each motion primitive generates or retrieves kinematic reference motion and executes its low-level policy to track the reference motion via physics simulation. For the locomotion primitives, we train a diffusion model on $\sim 25$ minutes of motion capture data to generate diverse walking motions to navigate through scenes, conditioned on waypoints. Simple tracking policies are then used to track the kinematic walking motion generated by the diffusion model. For the manipulation primitives, i.e. Pick-up and Put-down, we use a single reference trajectory and further rely on an environment-conditioned manipulation policy to achieve the needed variations of the motion. The quality of the motion can also be improved, e.g., making use of motion capture data of people picking up and putting down boxes can make the pick-up and put-down motion more natural.




\section{Locomotion}

\begin{figure}
    \centering
    \includegraphics[width=0.4\columnwidth]{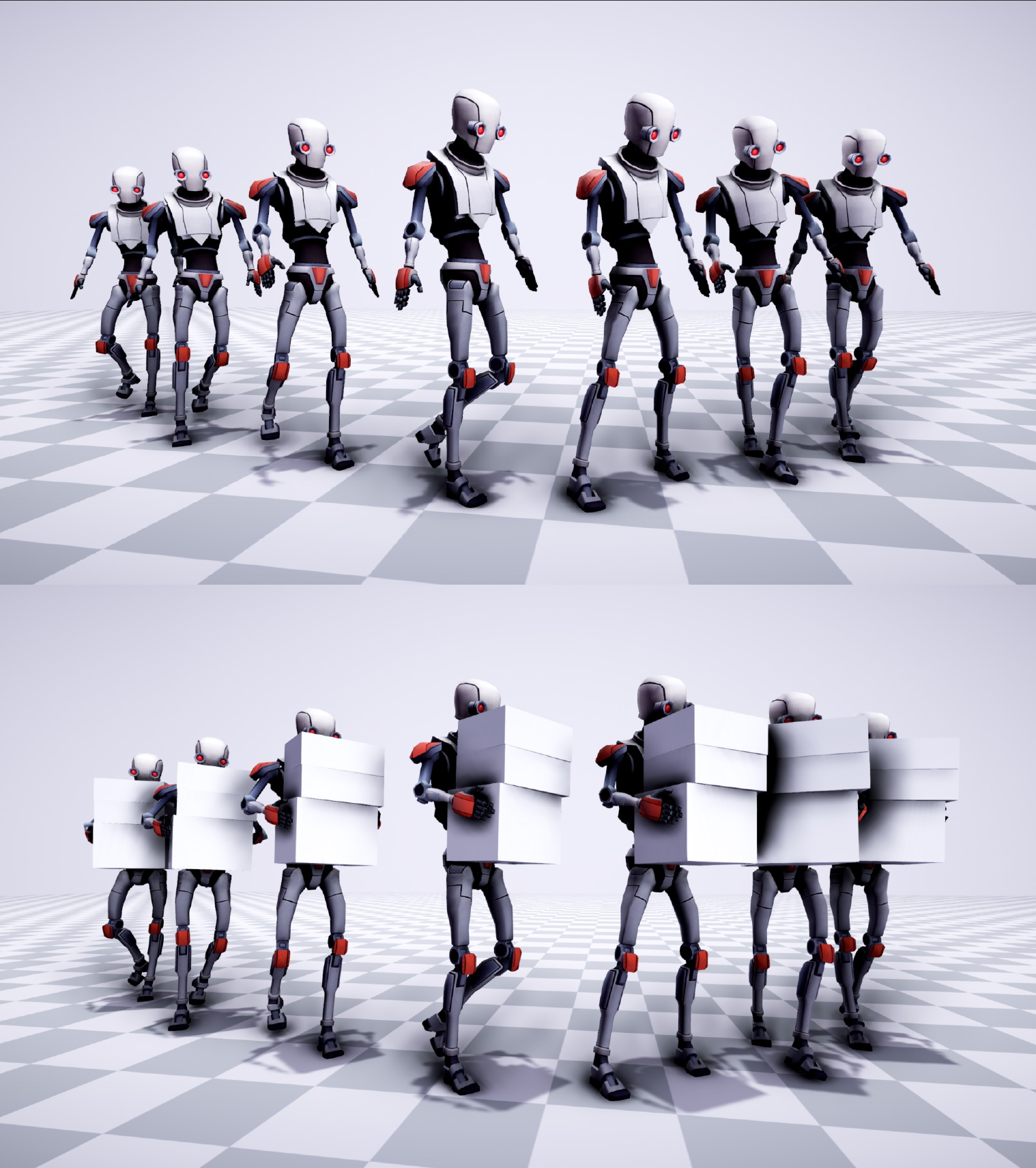}
    \caption{
    The two locomotion system primitives: walk-only (top) and walk-and-carry (bottom). 
    }
    \label{fig:loco}
\end{figure}

We develop two locomotion primitives: Walk-only and Walk-and-carry. They use the same diffusion model to generate kinematic whole-body reference trajectories given the waypoints, but have their own low-level RL policy to track the reference trajectory. Examples of the motions are shown in Fig.~\ref{fig:loco}. 


\subsection{Diffusion-based Whole-body Motion Generator}
We use a diffusion model \cite{diffusion} to generate whole-body navigation sequences. The diffusion model is trained with the locomotion dataset from \cite{2019-TOG-neural_state_machine}.
The diffusion model generates a full-body pose sequence from a noise sequence via an iterative denoising process. During training, noise is added to the data and a denoising model is then trained to reconstruct the original data from the noisy data. 

\subsubsection{Diffusion Process}
We follow the definition of diffusion from~\cite{2020-denoising-diffusion} as a Markov noising process with latents $\{\bm{z}_t\}_{t=0}^T$ that follow a forward noising process $q(\bm{z}_t | \bm{x})$, where $\bm{x} \sim p(\bm{x})$ is drawn from the data distribution.
The forward noising process is defined as
\begin{equation} \label{eq:1}
{
q(\bm{z}_{t} | \bm{x}) \sim \mathcal{N}(\sqrt{\bar{\alpha}_{t}}\bm{x}, (1 - \bar{\alpha}_{t})\bm{I}),
}
\end{equation}
where $\bar{\alpha}_{t} \in (0,1)$ are constants which follow a monotonically decreasing schedule such that when $\bar{\alpha}_{t}$ approaches 0, we can approximate $\bm{z}_{T} \sim \mathcal{N}(0,\bm{I})$.

In our setting with target trajectory conditioning $\bm{c}$, we reverse the forward diffusion process by learning to estimate $\bm{\hat{x}}_\theta(\bm{z}_t, t, \bm{c}) \approx \bm{x}$ with model parameters $\theta$ for all $t$.
We optimize $\theta$ with the ``simple'' objective introduced in~\cite{2020-denoising-diffusion}:
\begin{equation} \label{eq:2}
{
\mathcal{L}_\text{simple} = \mathbb{E}_{\bm{x}, t}\left[\| \bm{x} - \bm{\hat{x}}_\theta(\bm{z}_t, t, \bm{c})\|_2^2\right].
}
\end{equation}

\subsubsection{Kinematic Motion Representation}
Our diffusion model is conditioned by the waypoint vector $\bm{c} = \{\bm{c}_1^p, \bm{c}_2^p, \bm{c}_1^d, \bm{c}_2^d  \}$, which includes the target root position $\bm{c}^p \in \mathbb{R}^3$ and direction $\bm{c}^d \in \mathbb{R}^6$ for two goal waypoints given relative to the start state of the character. The direction is represented by the first two columns of the rotation matrix \cite{zhou2019continuity}.

We represent motions as sequences of poses $\bm{x} = \{ \bm{t}_i^p, \bm{t}_i^d, \tilde{\bm{t}}_i^p, \tilde{\bm{t}}_i^d, \bm{q}_i\}$ containing:
\begin{itemize}
    \item The Egocentric Root Trajectory: $\bm{t}_i^p \in \mathbb{R}^3, \bm{t}_i^d  \in \mathbb{R}^6$ are the predicted root position and direction at frame $i$ relative to the starting state.
    \item The Goal-Centric Root Trajectory: $\tilde{\bm{t}}_i^p \in \mathbb{R}^3, \tilde{\bm{t}}_i^d \in \mathbb{R}^6$ are the predicted root position and direction at frame $i$ relative to the next goal state.
    \item Predicted Joint Angles: $\bm{q}_i \in \mathbb{R}^{6M}$ is the predicted joint angles at frame $i$ for a skeleton of $M$ joints.
\end{itemize}

Given two waypoints, our model learns to synthesize kinematic motion sequences of $2N$ frames such that the first waypoint is reached at frame $N$ and the second is reached at frame $2N$. Thus, after the $N$-th frame, the Egocentric Root Trajectories are expressed relative to $\bm{c}_1$, and the Goal-Centric Trajectories are expressed relative to $\bm{c}_2$. 
{We choose $N = 15$ to produce waypoints that are spaced $0.5$ seconds apart.}

\subsubsection{Bidirectional Root Control}
In order to eliminate drift from generated trajectories, we adapt the bidirectional control scheme used in \cite{2019-TOG-neural_state_machine}, and represent the generated root position/orientation in both the frame of the start pose $\{\bm{t}_i^p, \bm{t}_i^d\}$ and the frame of the next waypoint $\{\tilde{\bm{t}}_i^p, \tilde{\bm{t}}_i^d\}$. At test time, the final trajectory is calculated as the weighted average of the two generated trajectories: $\lambda\{\bm{t}_i^p, \bm{t}_i^d\}+(1-\lambda)\{\tilde{\bm{t}}_i^p, \tilde{\bm{t}}_i^d\}$. We select $\lambda$ such that it linearly decreases to zero as the character approaches the waypoint over the course of $N$ frames. In practice, we find that this technique is sufficient to eliminate drift, while predicting only the egocentric trajectory causes discontinuous motions, as shown in the supplemental video.

\subsubsection{Model Architecture and Training}
\begin{figure}
    \centering
    \includegraphics[width=0.6\columnwidth]{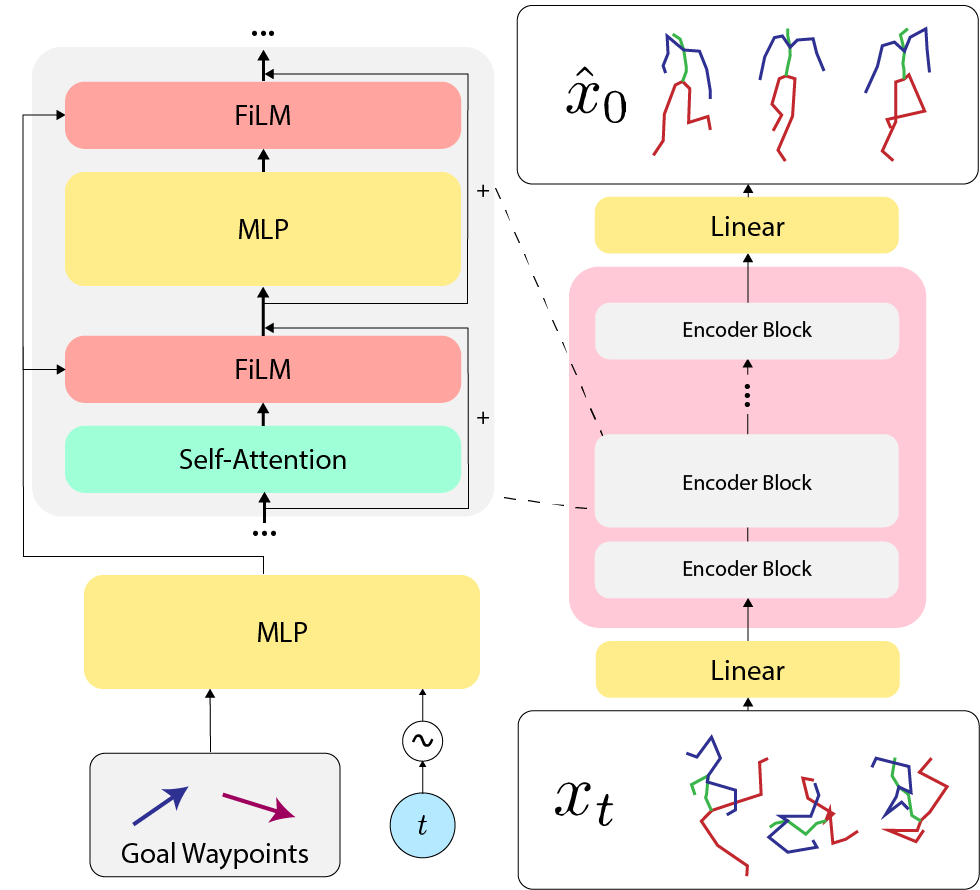}
    \caption{The diffusion model architecture.}
    \label{fig:diffusion_arch}
\end{figure}

The backbone of our diffusion model uses a transformer encoder architecture with feature-wise linear modulation (FiLM) \cite{2018-film-AAAI} units to inject conditioning information, as seen in Fig.~\ref{fig:diffusion_arch}. Training takes around 3 hours on a single RTX 3080 GPU.

\subsubsection{Parallel Long-form Generation}
\begin{figure}
    \centering
    \includegraphics[width=0.6\columnwidth]{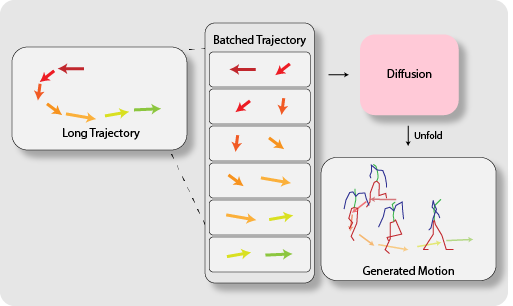}
    \caption{Enforcing continuity between short sequences is sufficient to create globally smooth motions.}
    \label{fig:diffusion_batch}
\end{figure}

The EDGE system~\cite{dance_diffusion} introduced the technique of \emph{parallel long-form generation} with diffusion models, a technique for generating batches of short, temporally consistent sequences such that they can be unfolded into an arbitrarily long sequence, noting that local consistency is sufficient to approach global consistency in many motion classes. We expand on this idea to solve the problem of tracking arbitrarily long trajectories. In this work, we represent a trajectory as a sequence of waypoints. We find that training a model to track as few as two waypoints at a time is sufficient to create globally consistent motions by enforcing consistency such that each sequence's first waypoint matches the second waypoint from the previous sequence (Fig.~\ref{fig:diffusion_batch}).

\subsection{Locomotion Control Policy}


Given the whole-body reference trajectory generated by the diffusion model, we use control policies to track the reference trajectory under physics constraints imposed by a physics simulator. We use reinforcement learning to train these policies to track the same motion data used to train the diffusion model. RAISIM~\cite{raisim} to simulate the rigid body dynamics and Proximal Policy Optimization~\cite{ppo} to optimize the policies. We simulate at \unit{120}{\hertz}, and query the policy at \unit{60}{\hertz}. Training requires 24 hours on a workstation with RTX A5000 GPU and a 12-core CPU.

\subsubsection{Policy Setup}
We represent our policy using a layer-wise mixture of experts neural network~\cite{2022-Soccer-Juggle}. The policy is tasked to track a reference motion that specifies the pose of the character over time. The input to the policy consists of the state of the character and the information about the current reference frame from the reference trajectory, namely:
\begin{itemize}
    \item The character: The input to the policy includes the current root height $p_z$, root linear and angular velocity $\bm{v}_{\text{root}}, \bm{\omega}_{{\text{root}}}$ in the local coordinate frame, and the joint orientations $\bm{q}$ and joint velocities $\dot{\bm{q}}$. 
    \item The reference frame: We compute the deviation of the root linear position, the root orientation, and the joint orientations between the next frame in the reference motion and the current character state, $\bar{\bm{t}}^p_{t+1} - \bm{t}^p$, $\bar{\bm{t}}^d_{t+1} \ominus \bm{t}^d$, and $\bar{\bm{q}}_{t+1} \ominus \bm{q}$, respectively. All vectors will be transformed into the local coordinate frame of the character. Unlike the diffusion model, here the root and joint orientations, $\bm{t}^d$ and $\bm{q}$, are represented as quaternions.
\end{itemize}

An imitation-based reward function is designed to encourage the policy to track the reference motion:
\begin{align*}
    &r_\text{joint} = \exp(-3\|(\bar{\bm{q}}_{t+1} \ominus \bm{q})\|^2), \\
   &r_\text{translation} = \exp(-\|\bar{\bm{t}}^p_{t+1} - \bm{t}^p\|^2),\\
   &r_\text{orientation} = \exp(-2\|\bar{\bm{t}}^d_{t+1} \ominus \bm{t}^d \|^2).
\end{align*}
Following \cite{2019-siggraph-scadiver}, we multiply the three reward terms so the policy receives a total reward of $r = r_\text{joint} \cdot r_\text{translation} \cdot r_\text{orientation}$ at each time step. Multiplicative reward function has the benefit of preventing the policy from ignoring any of the reward terms.


We adopt an action space in which the control of the character is generated by a combination of Proportional-Derivative (PD) control and torque control \cite{2022-Soccer-Juggle}. The torque applied in the simulation for each joint is generated via
\begin{align}
    \bm{\tau} = k_p (\bar{\bm{q}}_{t+1} \ominus \bm{q}) - k_d \dot{\bm{q}} + \bm{\tau}^a, \nonumber
\end{align}
where $\bm{\tau}^a$ is the action output from the policy to correct the naive PD feedback control. $k_p$ and $k_d$ are PD control gains and damping coefficients.

\zhaoming{We found that, without any regulation, $\bm{\tau}^a$ tends to fluctuate at a high frequency which causes ankles to vibrate, resulting in unrealistic foot sliding artifacts. To mitigate this issue, we apply a moving average of the policy output in simulation, i.e., the correction torque applied at time step $t+1$ will be calculated via $\bm{\tau}^a_{t+1} = 0.3\bm{\tau}^a_{t} + 0.7 \bm{\tau}^a$. We further introduce a reward term to encourage smoothness in policy output: 
\begin{align*}
    r_\text{smooth} = \exp(-\|(\bm{\tau}^a_{t+1} - \bm{\tau}^a_{t})\|^2).
\end{align*}
}

\subsubsection{Walk-only Policy}
Our control policy for locomotion is trained by imitating motion capture data of a human walking in various directions at various speeds. We use the same motion capture dataset \cite{2019-TOG-neural_state_machine} that trains the diffusion model to train the locomotion policies. At the beginning of each episode during training, a motion sequence is randomly sampled from the dataset, the character is initialized to the state of the first frame of the sampled sequence, and the reward is computed from the rollout. 

\subsubsection{Walk-and-carry Policy}
To generate carrying behavior during locomotion, we augment the locomotion dataset by fixing the arm to a pose that mimics object carrying. We also include the box position and box orientation in the character coordinate frame as input to the policy. During training, we initialize the box to be directly in front of the character between the two hands. In addition to the standard imitation-based reward, the policy also receives a reward for placing the hand on the opposing sides of the box and keeping the box upright:

\begin{align*}
&r_\text{box-hand} = \exp(-\| \bm{p}_{lhand} - \bm{p}_{lbox}\|^2 - \| \bm{p}_{rhand} -\bm{p}_{rbox}\|^2) \\
&r_\text{box-orientation} = \exp(-10\|(\bm{b}^d \ominus \bar{\bm{b}}^d)\|^2) \\
\end{align*}
where $\bm{p}_{lhand}$, $\bm{p}_{lbox}$, $\bm{p}_{rhand}$, $\bm{p}_{rbox}$ are predefined points on the hands and on the box in the world coordinate frame. $\bm{b}^d$ and $\bar{\bm{b}}^d$ are the current and the upright orientations of the box. 

The reward term for the box grabbed by the hands is defined as $r_\text{base} = r_\text{box-hand} + r_\text{box-orientation}$. When carrying multiple boxes in a stack, each box also needs to align with the box below:

\begin{equation}
r_\text{box-stack} = \exp(-\| \bm{b}^{p}_\text{top} - \bm{b}^p_\text{bottom}\|^2) \nonumber
\end{equation}
where $\bm{b}^p$ is the position of a box. During training, we randomize the number of boxes to carry for each episode. For each additional box $i$, we add another reward term $r_\text{box}^i$ which includes $r_\text{box-orientation}$ and $r_\text{box-stack}$ for box $i$. The total reward of walking and carrying $P$ boxes is the multiplication of all reward terms: $r = r_\text{joint} \cdot r_\text{translation} \cdot r_\text{orientation} \cdot  r_\text{base} \cdot \prod_{i=1}^{P-1} r_\text{box}^i + r_\text{smooth}$.

\section{Manipulation}
We design two manipulation primitives, Pick-up and Put-down. These two primitives share the same reference trajectory and the same low-level policy. 

\begin{figure}
    \centering
    \includegraphics[width=0.5\columnwidth]{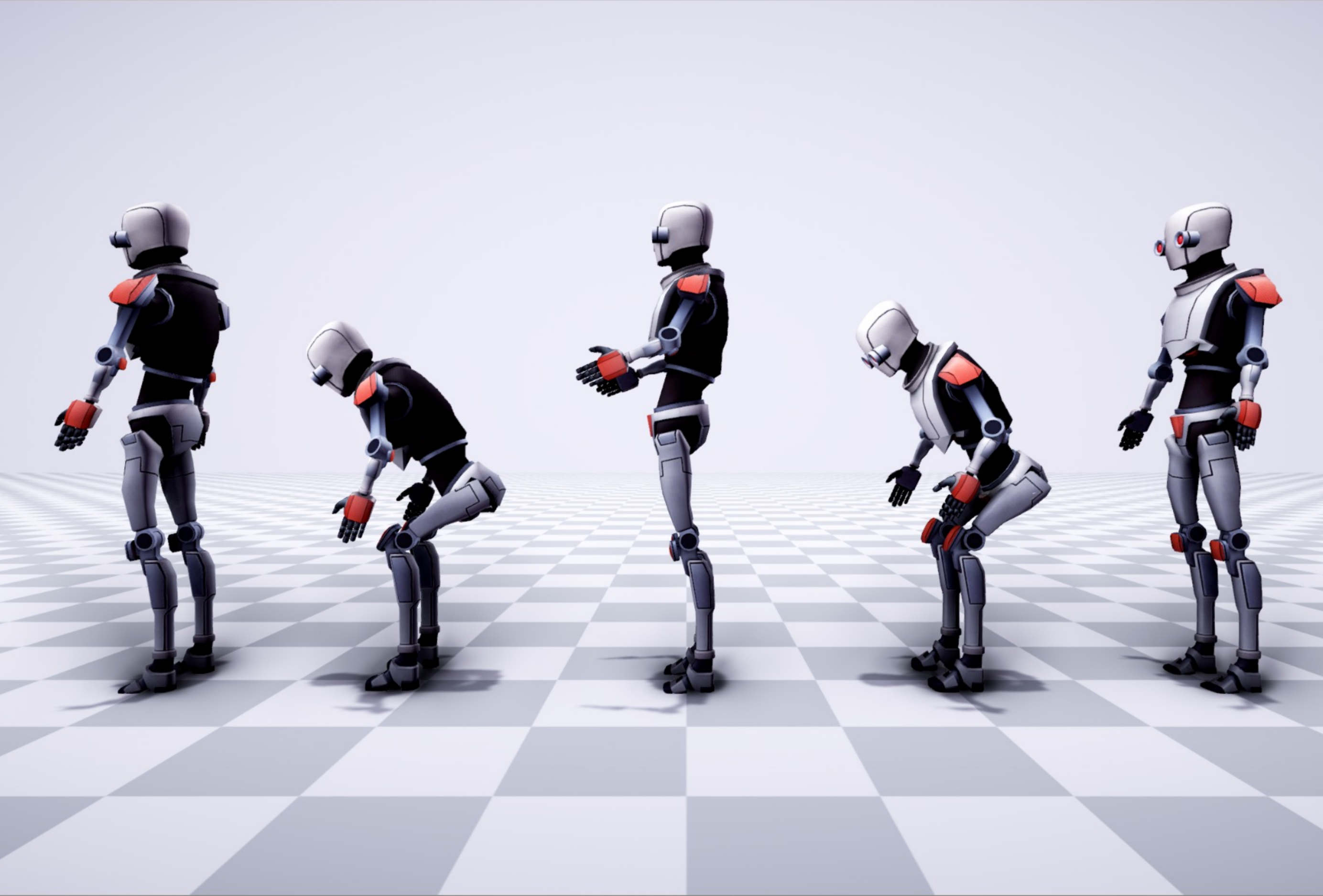}
    \caption{Reference motion for the pick up and put down motion.}
    \label{fig:pick_place_ref}
\end{figure}

\subsection{Manipulation Motion}
 A single reference trajectory is designed to train both primitives. The reference motion is shown in Fig~\ref{fig:pick_place_ref}. The trajectory consists of two stages: crouching down and standing up. We modify the arm poses to create simple pick-up and put-down motions. For pick-up, the arms linearly transition from a rest pose to a predefined carrying pose as the character crouching down, and stay in the carrying pose for the remaining of the motion. For put-down, the arms stay in the carrying pose when the character crouches down and linearly transition to the rest pose as the character stands up. 

\subsection{Manipulation Policy}
Although we only provide the manipulation policy a rough reference trajectory, we expect the RL policy to generalize to various scenarios such as different number of boxes with different sizes and weights being picked up and put down at different heights. Furthermore, we would like to have a unified policy for both pick-up and put-down. This allows us to concatenate the two tasks repeatedly to create natural and long-horizon animation. In addition, training both tasks simultaneously exposes the policy to a wide range of relative states, providing better generalization. During training, an episode starts with the box lying on a platform of random height in front of the character. The task for the policy is to pick up and put down the box on the platform repeatedly. Training the manipulation policies requires 24 hours on a workstation with RTX A5000 GPU and a 12-core CPU.

\subsubsection{Policy Setup}
The input to the manipulation policy includes, the position on the platform where the base box is placed, a one-hot vector to indicate whether the current task is to pick-up or put-down, in addition to the input to the locomotion policies. The reward function is the same as the one for walk-and-carry policy, with an additional reward term to guide the box to be at a certain height at every time step:
\begin{align}
    r_\text{box-height} = \exp\big(-10(\lambda h_\text{end} + (1-\lambda) h_\text{begin} - b^z_{base})^2\big), \nonumber
\end{align}
where $b^z_{base}$ is the height of the box being grabbed by the hands. The desired height is computed by interpolating between predefined height at the beginning and the end of the manipulation period, where the interpolation parameter $\lambda$ ranges from $0$ to $1$. The manipulation period for the put-down task starts from the beginning of the reference trajectory until the lowest crouch pose is reached, and the remaining of the trajectory is defined as the manipulation period for the pick-up task. For pick-up, $h_\text{begin}$ is the height of the platform and $h_\text{end}$ is defined at the torso height ($1.2m$ from the ground) of the character. The reverse is defined for the put-down task.

A total reward of $r = r_\text{joint} \cdot r_\text{translation} \cdot r_\text{orientation} + r_\text{base}$ is use at each time step. 
We do not multiply $r_\text{base}$ with the rest of the reward to give the policy more flexibility to trade off between achieving the manipulation goal and tracking the reference trajectory. We also train the policy to manipulate multiple boxes. In that case, the total reward function is defined as $r = r_\text{joint} \cdot r_\text{translation} \cdot r_\text{orientation} + (r_\text{base} \cdot \prod_{i=1}^{P-1} r^i_{box}) + r_\text{smooth}$.

\subsection{Transition Between Locomotion and Manipulation}
Our system allows for the sequential execution of locomotion and manipulation primitives. Since the manipulation policy is trained with an initial state of the character standing still in front of the boxes and the platform, it is essential to let the character comes to a full stop at the end of the locomotion to ensure a good transition to manipulation. Therefore, we simply add $30-60$ frames of standing still poses to be tracked by the locomotion policy. Transitions from manipulation to locomotion are more challenging due to the huge differences between the actions produced by the manipulation and the locomotion policy. We again add $30-60$ frames of standing pose at the end of manipulation and let both policies track the pose to produce two actions $\bm{a}_{manip}$ and $\bm{a}_{loco}$. The final action is a linear interpolation of the two: $(1-\lambda) \bm{a}_{manip} + \lambda \bm{a}_{loco}$, where $\lambda \in \mathbb{R}$ linearly progresses from $0$ to $1$ during the transition period (the added $30-60$ frames).
\section{Planner}



\begin{figure}
    \centering
    \includegraphics[width=0.5\columnwidth]{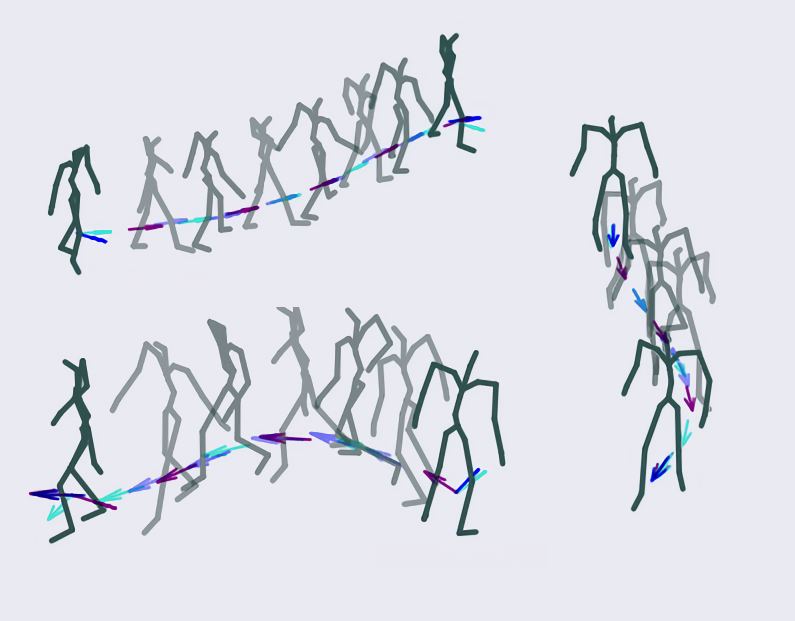}
    \caption{Trajectories are generated by an A* planner, and tracked by a diffusion model.}
    \label{fig:plan}
\end{figure}

We use a kinodynamic A* planner to automatically avoid in-scene obstacles and generate paths between boxes using a single waypoint for each box location and desired pick-up orientation, while maintaining realistic human navigation velocities and accelerations.
Each planned trajectory is smoothed with cubic splines to maintain reasonable linear and angular velocities, and goal conditioning for the diffusion model is then sampled from the smoothed path.
The choice of the A* algorithm is arbitrary, and alternatives, (i.e. D* or RRT) could be easily substituted out-of-the-box. 
An example of such a planned trajectory and the resulting whole-body motion generated by the diffusion model is shown in Fig.~\ref{fig:plan}.

\begin{figure}
    \centering
    \includegraphics[width=0.5\columnwidth]{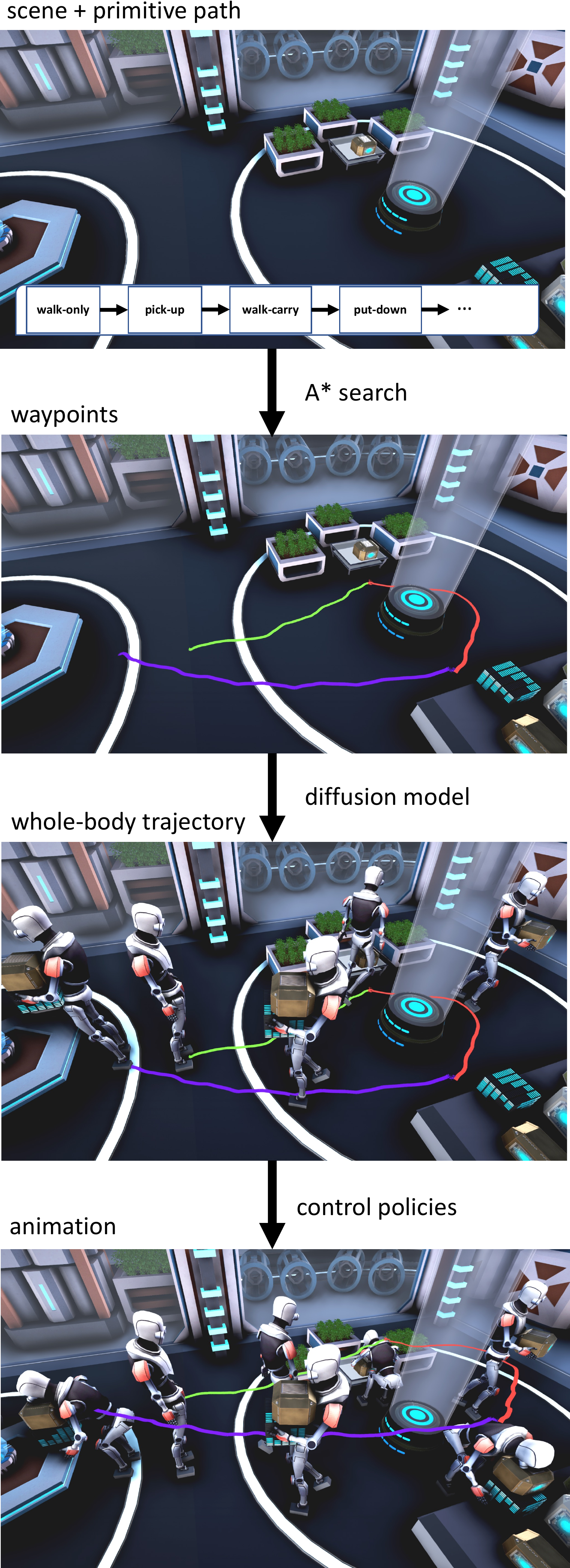}
    \caption{Our system can solve scene arrangement tasks where the character moves boxes around to a goal configuration.}
    \label{fig:scene1}
\end{figure}

\begin{figure}
    \centering
    \includegraphics[width=0.5\columnwidth]{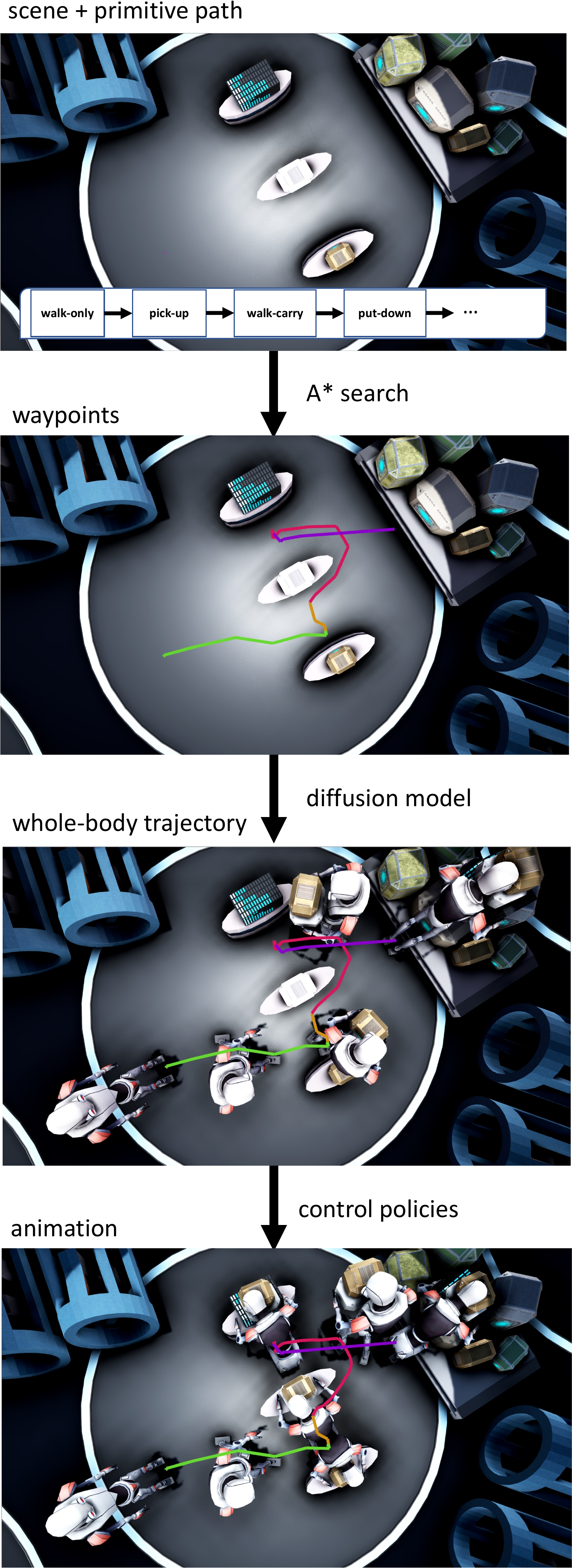}
    \caption{We demonstrate a task that involves rearrangement of 3 boxes.}
    \label{fig:scene2}
\end{figure}
\section{Results and Discussion}
We evaluate our method by generating diverse animations of a character performing box arrangement tasks in a contextual 3D scene. We also demonstrate the generalizability and robustness of our method. Finally, we show an ablation study to validate the effect of the bidirectional root control used in the diffusion model for locomotion.

We model the character as an articulated rigid body system, adapted from Scadiver \cite{2019-siggraph-scadiver}. It has $59$ degrees of freedom, including a floating base. We use RAISIM~\cite{raisim} as the physics simulator. The character is controlled via Stable proportional-derivative controllers~\cite{2011-StablePD}. 

\subsection{Diversity of Rearranging Tasks}
To test our pipeline end-to-end, we generated two long-horizon animations of the character rearranging assorted boxes in two different scenes. The qualitative evaluation shows that diverse animations can be easily simulated using the four motion primitives and their associated policies. Fig.~\ref{fig:scene1} and Fig.~\ref{fig:scene2} demonstrate how we use our system to solve the tasks. 

\zhaoming{We further create two scenarios to demonstrate other interesting animation scenarios using our system, namely rearranging cylinder objects and playing the Tower of Hanoi game (Fig.~\ref{fig:cylinder_tower}). The full animations can be viewed in the supplemental video.}

\begin{figure*}
    \centering
    \includegraphics[width=0.8\columnwidth]{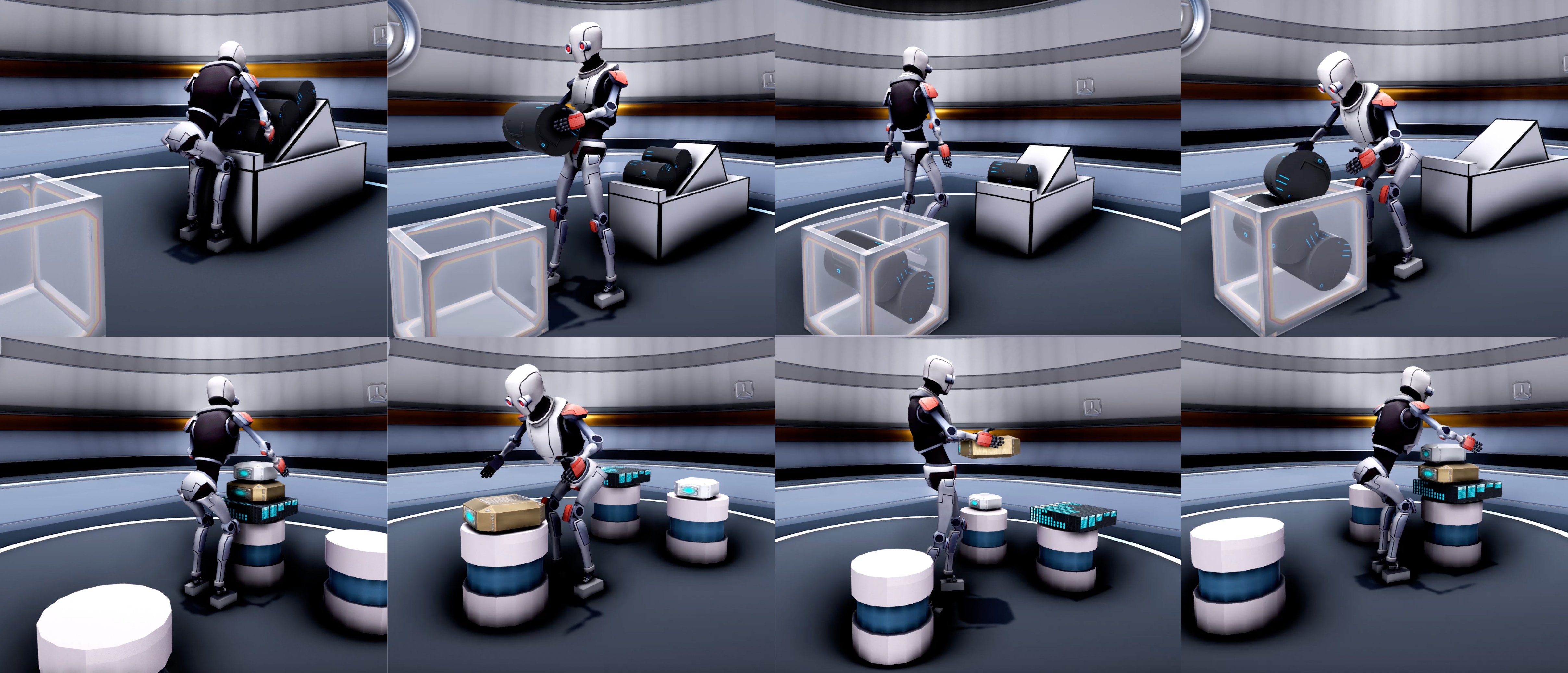}
    \caption{Our system can generate animations such as moving cylinder objects (TOP) or playing Tower of Hanoi (BOTTOM).}
    \label{fig:cylinder_tower}
\end{figure*}


\subsection{Generalizability of Manipulation Policy}
During training, we varied the initial position of the platform, the height of the platform, and the width of the boxes to generate more robust policies. We tested our manipulation policy with different scenarios, including robustness to variation in pick-up and placing location, the weight of the boxes, and the box width. Table~\ref{tab:box_test} shows the range of each varying parameter within which our policy can successfully complete the manipulation tasks.


\begin{table}
    \centering
    \footnotesize
    \caption{We test the robustness of the pick-up and put-down policy in various scenarios. The default position of the platform to pick up and put down the boxes is $(0.4, 0)$ meters away from the character in the frontal plane, with a height of $0.5$ meters. By default, each box weighs $1$ kg with a width of 0.4 $meter$. We vary each parameter one at a time while keeping other parameters fixed and record the extreme values for which the policy can pick up and put down the boxes at least $3$ times. We also record the training distribution.} 
    \begin{tabular}{|P{0.12\textwidth}|P{0.12\textwidth}|P{0.12\textwidth}|P{0.12\textwidth}|P{0.15\textwidth}| P{0.12\textwidth}| }
        \hline
         & sagittal & frontal & transverse & weight (per box) & width \\
         \hline
         training & 0.35, 0.40 & -0.10, 0.10 & 0.3, 0.7 & 1.0 & 0.3, 0.5\\
         \hline
         one & 0.20, 0.68 & -0.30, 0.32 & 0.05, 1.00 & 8.0 & 0.12, 0.79 \\
         \hline
         two & 0.28, 0.70& -0.29, 0.29& 0.09, 0.92 & 3.7 & 0.14, 0.79\\
         \hline
         three & 0.30, 0.52 & -0.26, 0.28 & 0.13, 0.92 & 2.5 & 0.16, 0.79\\
         \hline
    \end{tabular}
    \label{tab:box_test}
\end{table}

\subsection{Robustness of Locomotion}
The combination of the diffusion model and the locomotion policies is robust to simulate a wide range of scenarios. 
To test how well the diffusion model can work with the RL policies, we generate $10$ trajectories of length $10$ seconds using random waypoints with the diffusion model. Our RL policy is able to track all the trajectories successfully even though the RL policy and the diffusion model were trained independently.

We also show that the walk-and-carry policy is robust to withstand external perturbation--the character was able to hold on to all the boxes while being hit by heavy projectiles. We also test the robustness of the policy to the width and weight of the boxes being carried. Statistics of the tests are recorded in Table~\ref{tab:carry_robustness}.

\begin{table}
    \centering
    \footnotesize
    \caption{We test the robustness of the carrying policy by varying the box width and weight. We also test the robustness by throwing projectiles from behind with a horizontal velocity of $3 m/s$ every $0.5$ second for $10$ seconds and record the maximum projectile weight the policy can withstand.}
    \begin{tabular}{|P{0.2\textwidth}|P{0.2\textwidth}|P{0.2\textwidth}| P{0.2\textwidth} |}
        \hline
         & weight (per box) & width & projectile weight  \\
         \hline
         training & 1.0 & 0.3, 0.5 & 0\\
         \hline
         one &  5.4 & 0.21, 0.56  &  3.7\\
         \hline
         two &  2.1 & 0.23, 0.56 & 2.6 \\
         \hline
         three &  1.1 & 0.31, 0.60 & 0.9\\
         \hline
    \end{tabular}
    \label{tab:carry_robustness}
\end{table}

\subsection{Comparison To DReCon}
We use a diffusion model to generate kinematic trajectory for our low-level control policy to track. There are many other ways to generate such trajectory, e.g., motion matching~\cite{2018-motion_matching}, as is used in DReCon~\cite{2019-siggraph-drecon}. Here we compare how well our low-level controller can track the trajectories generated via the diffusion model and motion matching. For the diffusion model, we generate waypoints such that the first two waypoints are at the same point and the next $20$ waypoints are spaced 0.4 meters apart so the average velocity is $0.8 m/s$ for $10$ seconds. For motion matching, we use the root velocity and orientation of the trajectory as features for motion query and generate a motion that has $0$ velocity at the first second and has $0.8 m/s$ for the next 10 seconds. Table~\ref{tab:drecon_comparison} record the average performance of the low-level policy in terms of tracking the joint, root position and root orientation. Tracking performance is higher for the diffusion model, even though the low-level policy is never trained with the diffusion model generated data. This indicates our diffusion model generates data that is more in distribution with the motion capture data. DRecon use the output of the motion matching to train the control policy, which can explain their better performance in terms of tracking the motion matching output. Furthermore, motion matching requires storing the motion database in memory for query, which becomes unscalable as we would want to track more diverse motion in the future. On the other hand, diffusion model is able to condense large motion dataset into a neural network, as demonstrated by recent work, e.g.,~\cite{dance_diffusion, 2022-motion_diffusion}.

\begin{table}
    \centering
    \footnotesize
    \caption{We test the performance of low-level control policy in terms of tracking reward, with reference motion generated by our diffusion model and motion matching.}
   \begin{tabular}{|P{0.2\textwidth}|P{0.2\textwidth}|P{0.2\textwidth}| P{0.2\textwidth} |}
        \hline
         & $r_\text{joint}$ & $r_\text{translation}$ &  $r_\text{orientation}$  \\
         \hline
         diffusion & 0.94 & 0.92 & 0.98\\
         \hline
         motion matching &  0.90 & 0.62  &  0.95\\
         \hline
    \end{tabular}
    \label{tab:drecon_comparison}
\end{table}

\subsection{Ablation on Bidirectional Root Control}
\begin{figure}
    \centering
    \includegraphics[width=0.4\columnwidth]{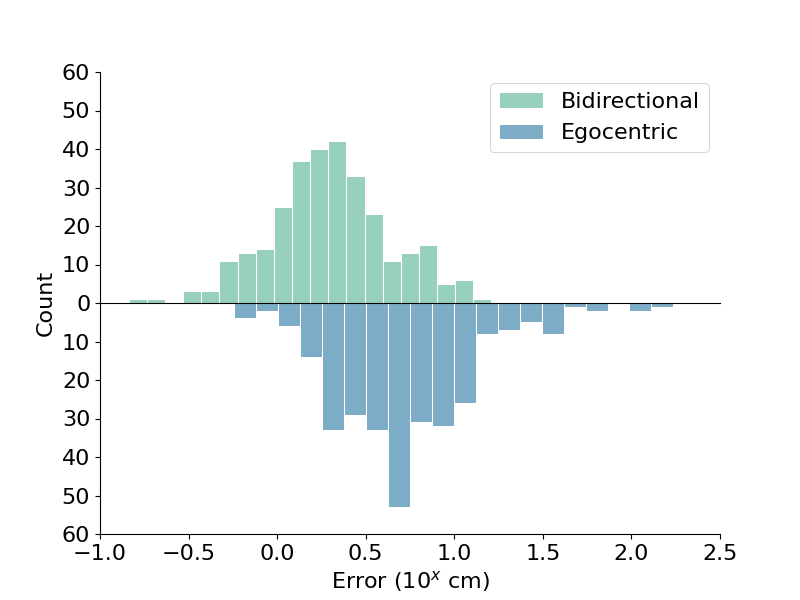}
    \caption{Errors of the bidirectional and egocentric models, compared over different reference trajectories ($n = 297$). 
    The bidirectional model performs consistently better than the egocentric model, which also shows poor long-tail behavior with significant deviation in out-of-distribution scenarios.}
    \label{fig:bidirectional_ablation}
\end{figure}

We validate the bidirectional control model qualitatively and quantitatively through an ablation study, comparing it to a model that only uses the predicted egocentric trajectory. We find that although in simpler cases, both models perform similarly on a qualitative level, challenging scenarios involving more precise turning motions cause the egocentric model to visibly deviate from the path, while the bidirectional model does not face these issues.
We plot the average error from reference trajectory across both models in Figure~\ref{fig:bidirectional_ablation}. The bidirectional model reduces error by 70\% on average and is more stable than its egocentric counterpart, which has poor long-tail behavior.
For qualitative examples, see our supplemental video.

\zhaoming{
\subsection{Comparison to Other Hierarchical Framework.}
Different designs of hierarchical frameworks have been proposed previously. Notably, Catch and Carry\cite{2020-catch_and_carry} and Adversarial Skill Embedding (ASE) \cite{2022-TOG-ASE} train a low-level control policies to track a wide range of motion capture data (similar to ours) and a high-level policy to control in a learned latent space to accomplish tasks. 
The latent high-level action space used by Catch and Carry can cause inefficient high-level policy training, which can take weeks to train. ASE employs a smart design of the high-level action space, which can decrease the learning efficiency of the low-level control policies. For example, in ASE, training the low-level policy takes $10$ days with a much faster simulation engine, while our policy can be trained in one day. It is also not clear how to learn low-level skills that can interact with objects within the ASE framework. Our system allows for efficient training of both low-level policies and mid-level motion generators and our high-level planner allows for the rapid design of new tasks.
}

Our system also has limitations compared to the two-level hierarchical framework. For example, there is no feedback between the mid-level motion generator and the low-level controller, which can lead to less robust behaviors under perturbation. We evaluate how root tracking errors can affect the performance of the policy. 
Even though our locomotion controller can achieve high tracking reward for tracking the global root position, there are still tracking errors that can cause task failure. For example, in a cluttered environment, a slight deviation from the planned root trajectory can cause collisions with the environment. Furthermore, if the character fails to arrive at the planned location to execute the pick-up primitive or put-down primitive, it may fail to accomplish the task. We examine the tracking accuracy of the locomotion controller by tracking a straight walking motion that lasts for 10 seconds with an average speed of $0.8 m/s$. The final root position has a maximum $0.5$ meter error in the forward direction and $0.15$ meter error in the lateral direction.
Since the policy can observe the tracking error and constantly try to catch up or slow down, it can exhibit artifacts such as stumbling behaviors and sometimes bump into the environment (see video for some of these artifacts).
Control policy with high root tracking accuracy is essential for completing tasks that span longer time or in a more cluttered environment.

In addition, our high-level planner requires a predefined primitive path, which can generate less natural transition between primitives. For example, our character needs to assume a standing pose before transitioning from a locomotion primitive to a manipulation primitive. It is also hard to generate a natural stepping sequence using the A* planner. For example, generating side stepping behaviors when getting close to the target location, or generating backward steps to avoid collision with the platform after picking up an object.

\section{Conclusions}


We demonstrate physically simulated characters that can fetch, carry, place and stack objects in their environments.
These types of abilities to interact with the environment are critical for fully realizing
physics-based worlds with everyday human behaviors, whether in simulation or in robotics.
We also show that diffusion models can be used as an effective mid-level planner -- they provide motions of sufficient quality
that their output can be directly tracked with RL-based online imitation controllers.
We further demonstrate how to add object interactions to low-level RL-driven physics-based imitation methods.


Our work has a number of limitations that also point to future work.
Our system currently consists of a number of discrete controllers.
A future avenue for exploration could investigate developing a single diffusion model that 
generates the mid-level motion plan for all phases of all tasks.
This could then be used in conjunction with a single low-level controller
that is also capable of all the phases of loco-manipulation tasks. With a unified model, our current limitation during transitions between manipulation and locomotion can be resolved.
Dynamic replanning with the diffusion model would be
a natural approach to making the behaviours robust to unexpected events.
Obstacle avoidance can also potentially be incorporated into the diffusion model.
We wish to be able to carry and manipulate heavy boxes.  This is likely to require
hand models with fingers that can wrap around the bottom of the box or to grab box handles.
Carrying and moving heavier objects is also likely to require 
learning to use the full body to provide support while carrying them.
Lastly, additional fidelity to human motion may be achieved via
modeling of gaze, biomechanical models, and controllers that learn richer manipulation.

\begin{acks}
We thank Takara Truong for helping create the reference motion for the pick-up and put-down policy and helpful discussion.
\end{acks}

\bibliographystyle{ACM-Reference-Format}
\bibliography{references}




\end{document}